\documentclass[10pt,twocolumn,letterpaper]{article}

\usepackage{cvpr}              

\usepackage{colortbl}
\usepackage{pifont}
\usepackage[accsupp]{axessibility} 

\definecolor{cvprblue}{rgb}{0.21,0.49,0.74}
\usepackage[pagebackref,breaklinks,colorlinks,allcolors=cvprblue]{hyperref}

\def\paperID{ } 
\def\confName{CVPR}
\def\confYear{2026}

\title{HAM: A Training-Free Style Transfer Approach via Heterogeneous Attention Modulation for Diffusion Models}

\author{
    {
        Yeqi He$^{1,2}$\thanks{This work is done during the intern in VIPL group, ICT, CAS.}$\phantom{0}$ 
        Liang Li$^{2}$\thanks{Corresponding author}$\phantom{0}$
        Zhiwen Yang$^{1,2}$\footnotemark[1]$\phantom{0}$
        Xichun Sheng$^{3}$$\phantom{0}$
        Zhidong Zhao$^{1}$$\phantom{0}$ 
        Chenggang Yan$^{1}$$\phantom{0}$
    }\\
    {\small $^1$ Hangzhou Dianzi University~}
    {\small $^2$ Institute of Computing Technology, Chinese Academy of Sciences~}
    {\small $^3$ Macao Polytechnic University}\\
    {\tt\small 
        \{yeqihe,zhiwen.yang,zhaozd,cgyan\}@hdu.edu.cn$\phantom{0}$
        liang.li@ict.ac.cn$\phantom{0}$
        p2314922@mpu.edu.mo
    } \\
}

\begin{document}
\maketitle
\begin{abstract}
\label{sec0:abstract}

Diffusion models have demonstrated remarkable performance in image generation, particularly within the domain of style transfer. 
Prevailing style transfer approaches typically leverage pre-trained diffusion models' robust feature extraction capabilities alongside external modular control pathways to explicitly impose style guidance signals.
However, these methods often fail to capture complex style reference or retain the identity of user-provided content images, thus falling into the trap of style-content balance.
Thus, we propose a training-free style transfer approach via \textbf{h}eterogeneous \textbf{a}ttention \textbf{m}odulation (\textbf{HAM}) to protect identity information during image/text-guided style reference transfer, thereby addressing the style-content trade-off challenge.
Specifically, we first introduces style noise initialization to initialize latent noise for diffusion. 
Then, during the diffusion process, it innovatively employs HAM for different attention mechanisms, including Global Attention Regulation (GAR) and Local Attention Transplantation (LAT), which better preserving the details of the content image while capturing complex style references.
Our approach is validated through a series of qualitative and quantitative experiments, achieving state-of-the-art performance on multiple quantitative metrics.

\end{abstract}

%

\begin{figure}[t]
    \centering
    \begin{subfigure}[b]{1\columnwidth}
        \centering
        \includegraphics[width=\columnwidth, page=1]{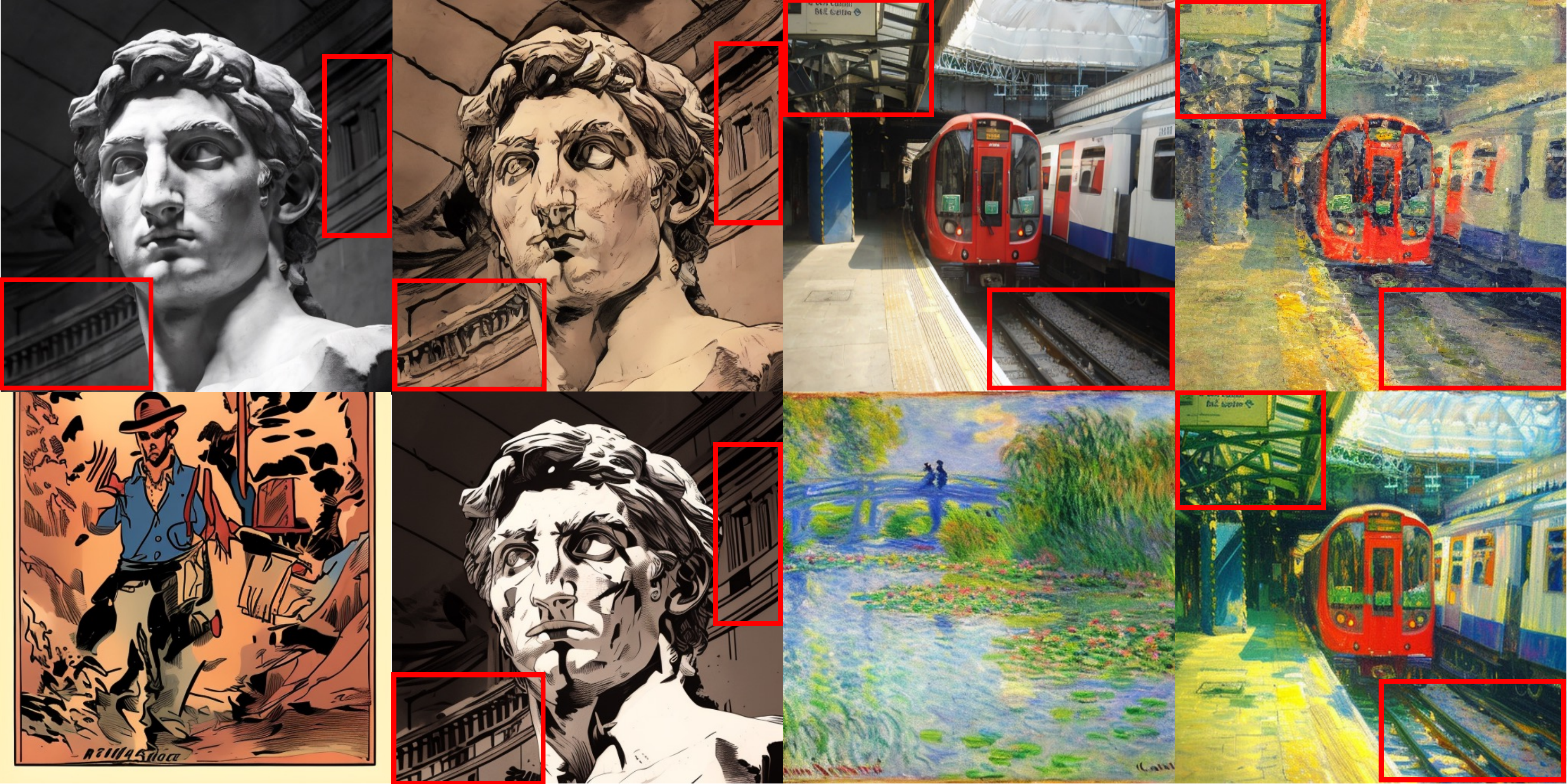}
        \caption{Image-guided style transfer results. Compare with StyleID~\cite{styleid}.}
        \label{fig:intro_page1}
    \end{subfigure}
        
    \begin{subfigure}[b]{1\columnwidth}
        \centering
        \includegraphics[width=\columnwidth, page=2]{image/intro.pdf}
        \caption{Text-guided style transfer results. Compare with DiffArtist~\cite{artist}.}
        \label{fig:intro_page2}
    \end{subfigure}
    \caption{Comparative results of style transfer methods: 
         Content image (top-left), style reference (bottom-left), 
         baseline method (top-right), and our HAM (bottom-right). 
         Red boxes denote significant identity retention disparities.}
    \label{fig:intro}
\end{figure}
\section{Introduction}
\label{sec1:intro}

The development of generative diffusion models has propelled advances in text-to-image generation~\cite{dalle,dalle2,dalle3,sd2,sdxl,sd3}, image editing~\cite{Tumanyan_2023_CVPR,Kawar_2023_CVPR,Liu_2024_CVPR}, and related fields.
Among these, the generative diffusion models have also been applied to style transfer~\cite{Huang_2017_ICCV,Zhang_2023_CVPR}, specifically migrating the style references of a given content image to a designated style preset while preserving its identity information.
Given the powerful text-to-image generation capabilities of generative diffusion models, they have brought a efficient yet challenging new paradigm to style transfer.

Diffusion-based style transfer approaches typically utilize the inherent generative capabilities of pre-trained diffusion models~\cite{sd2,sdxl,sd3} to achieve style migration for given content images.
Some approachs, involves explicit style-content feature decoupling~\cite{kwon2023diffusionbasedimagetranslationusing,frenkel2024implicit,wang2024instantstyle,xing2024csgo}, leveraging interpretability to process style features while retaining content features, and incorporates style attributes into pre-trained diffusion models via LoRA~\cite{lora} or ControlNet~\cite{controlnet} for fine-tuning-based style control.
While effective for style transfer, these methods are computationally intensive and lack robustness, as their performance on diverse style references is highly sensitive to the extent of fine-tuning.


To fully utilize the powerful generative capabilities of diffusion models, training-free style transfer methods have been proposed. 
Representative works like StyleID~\cite{styleid} and DiffArtist~\cite{artist} achieve stylization by injecting the keys and values from style features, extracted via diffusion model inversion, into the self-attention layers during generation.
As theory indicates~\cite{prompt,survey}, self-attention features, including queries, keys, and values, collectively encode various semantic and spatial relationships. 
Consequently, such methods that solely rely on self-attention manipulation results in insufficient style or distorted content, as Fig.~\ref{fig:intro} demonstrates, resulting in an imbalance between style and content.

In this work, we propose a training-free style transfer approach via \textbf{h}eterogeneous \textbf{a}ttention \textbf{m}odulation for diffusion models (\textbf{HAM}), which significantly improves the style-content balance capability of style transfer.
Our method utilizes a style/content teacher model obtained from style references and content images, and then uses HAM to combine and share the knowledge from the teacher model  to the student generator, thereby achieving style transfer.

Building upon framework process, we propose a \textbf{s}tyle-\textbf{i}nfused \textbf{n}oise \textbf{i}nitialization (\textbf{SINI}) at timestep $T$, where the initial latent noise is derived by fusing the inverted noise from the reference style and content images through adaptive instance normalization.
To better preserve identity information, the fused initial latent noise is then modulated by the inverted content initial latent noise.
Subsequently, in the process of generating and diffusing stylized images, we introduce our HAM, comprising the \textbf{g}lobal \textbf{a}ttention \textbf{r}egulation (\textbf{GAR}) and \textbf{l}ocal \textbf{a}ttention \textbf{t}ransplantation (\textbf{LAT}), respectively. 
GAR is a mechanism designed to preserve content and introduce style by exerting macroscopic control over attention injection, thereby maintaining the original spatial and style semantic structure. 
Specifically, it fuses content and style teacher's attention projections into the specific projections, ensuring their statistical distribution aligns with the corresponding attention projections in the student generator. 
These specific projections are then reconciled with the ones to further stabilize content/style information.
After that, to implement precise style/content control in the absence of text prompts, we introduce LAT through feature operations in cross-modal cross-attention. 
To better preserve identity information, we inject and weight the query from the content teacher into the query from student generator, while the original key/value pairs are replaced directly with those from the style teacher to ensure effective style guidance.
This heterogeneous attention modulation effectively decouples style guidance and content preservation, enabling high-fidelity stylization without compromising structural integrity.

Our GAR and LAT adapt to base model architectures: operating within self-/cross-attention for SD2.1~\cite{sd2}, and joint-/dual-attention for SD3.5~\cite{sd3}, respectively.

The main contributions are summarized as follows:
\begin{itemize}
    \item We propose a training-free stylized image generation method, HAM, which can achieve high-quality stylized image generation without the need for gradient optimization of style images.    
    \item In our proposed HAM, GAR effectively macroscopically introduces features from the style/content teacher into the student generator, while LAT precisely controls the guidance between style and content. The combined effect improves the quality of the generated stylized images.
    \item We demonstrate HAM's universal compatibility across DDIM-based (SD2.1) and DiT-based (SD3.5) architectures, achieving state-of-the-art performance on multiple metrics through comprehensive evaluations.
\end{itemize}
\section{Related Work}
\label{sec2:related}

\subsection{Text-Driven Image Generation}
\label{sec2.1:text-driven image generation}

With the advancement of deep learning~\cite{10607969,cui2025debiased,cui2024stochastic}, Text-driven image generation~\cite{dalle,dalle2,dalle3} has enabled the synthesis of highly realistic and semantically coherent images.
Advances in text encoder architectures, exemplified by the SD series~\cite{sd2,sdxl,sd3}, have significantly improved text-to-image synthesis through structural refinements and systematic optimization.
These technical developments yield quantifiable gains in output visual quality, particularly in enhanced texture detail and resolution fidelity, while reinforcing model robustness in maintaining precise semantic alignment with complex, compositional textual prompts.
The progress in generative foundations~\cite{peng2025survey,ll5} has also propelled developments in related areas, including: (1) text-guided image editing~\cite{survey,lv2024pick,mo2024freecontrol}, (2) semantic-aware style transfer~\cite{Huang_2017_ICCV,Zhang_2023_CVPR}, and (3) emerging multimodal generative applications beyond static imagery~\cite{zhao2025heterogeneous,zhang2025prosody,11121533,zhao2026temporal,10433795,zhang2024speaker2dubber}.

\subsection{Image Style Transfer}
\label{sec2.2:image style transfer}

Style transfer aims to apply a reference style's visual characteristics to a content image while preserving its structural and semantic core, hinging on disentangling content and style representations. 
Current approaches fall into tuning-based and training-free categories. Tuning-based methods adapt models through parameter updates (e.g., coupling style-specific content or adding lightweight adapters), exemplified by ControlNet~\cite{controlnet} training conditional copies, B-LoRA~\cite{frenkel2024implicit} innovating weight optimization, and CSGO~\cite{xing2024csgo} using curated adapters. 
Training-free methods manipulate diffusion mechanisms during inference without parameter changes, altering attention maps/activations to redirect synthesis—e.g., P2P~\cite{prompt} injecting reconstructed cross-attention maps and StyleID~\cite{styleid} fusing features into self-attention layers for zero-shot stylization.
\section{Method}
\label{sec3:method}

In this section, we present our proposed method comprising three core modules: global attention regulation, local attention transplantation, and style-infused noise initialization. 
Compatible with both DDIM-based SD2.1~\cite{sd2} and DiT-based SD3.5~\cite{sd3} architectures, our method is detailed in the following subsections, with technical descriptions primarily based on the SD2.1 framework.

\begin{figure*}[t]
\centering
\includegraphics[width=1\textwidth]{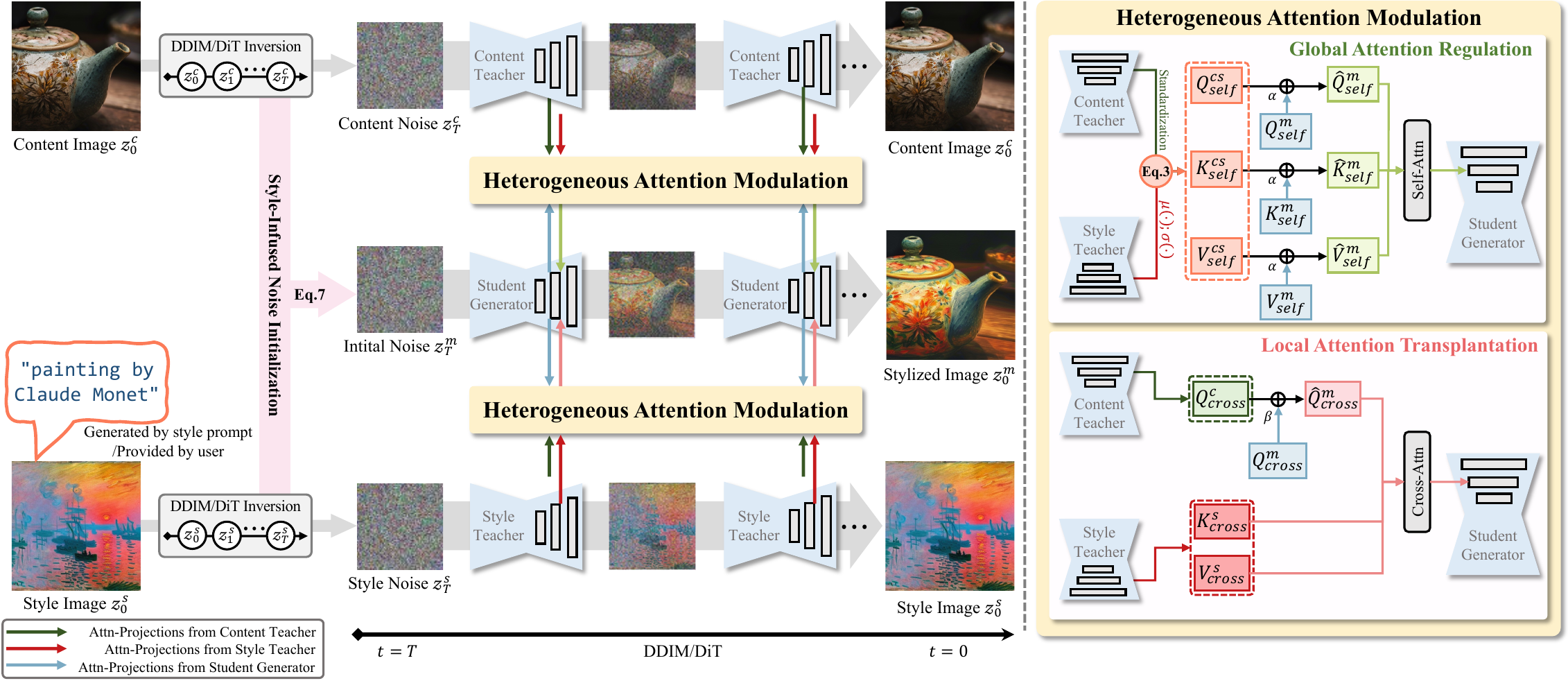}
\caption{The overall pipeline of our method.
Our proposed method consists of three main modules: global attention modulation, local attention transfer, and style injection noise initialization, which act on the self-attention, cross-attention, and noise initialization stages respectively.
Through the joint modulation of the three modules, the final stylized image can retain more content identity information and capture and transfer complex style references.
}
\label{fig:main}
\end{figure*}

\subsection{Preliminaries}
\label{sec3.1:preliminaries}

Before delving into the specifics of our proposed methodology, we initially provide a comprehensive background on the fundamental techniques that underpin our method.

\subsubsection{Latent Diffusion Models}
\label{sec3.1.1:latent diffusion models}

Latent Diffusion Models (LDMs)~\cite{sd2} represent a prominent image generation framework that maps images into the latent space of the Variational Auto-Encoder (VAE)~\cite{vae} and subsequently leverages the powerful generative capacity of diffusion models to synthesize high-quality images while optimizing computational efficiency. 
The most representative class of such methods comprises the Stable Diffusion (SD)~\cite{sd2,sdxl,sd3}, which utilizes text prompts as input conditions to denoise latent noise for generating high-quality, high-fidelity images. 
The overall denoising process is formally described by Eq.~\ref{eq:ldm}.

\begin{equation}
\begin{aligned}
\mathcal{L}_{\theta}=\mathbb{E}_{z,t,c,\epsilon\in\mathcal{N}(0,1)}\left[\left\|\epsilon-\epsilon_{\theta}(z_t,t,c)\right\|^2_2\right],
\end{aligned}
\label{eq:ldm}
\end{equation}
where $t$ denotes the current diffusion time step, $z_t$ represents the latent noise vector corresponding to time step $t$, $c$ signifies the conditioning text prompt, $\epsilon$ is the Gaussian-distributed noise sampled from $\mathcal{N}(0,1)$, and $\epsilon_{\theta}(z_t,t,c)$ denotes the noise component predicted by the model.

\subsubsection{Attention Mechanism}
\label{sec3.1.2:attention mechanism}

To enhance the quality of synthesized images and effectively integrate external conditioning information, Stable Diffusion incorporates multiple groups of self-attention and cross-attention blocks~\cite{attention}, which are typically arranged in complementary pairs. 
As previously established, our proposed methodology focuses on modulating this dual-attention mechanism, specifically by concurrently targeting both self-attention and cross-attention blocks. 
The precise mathematical formulations governing these two attention operations are formally expressed in Eq.~\ref{eq:attention}.

\begin{equation}
\begin{aligned}
\text{Attention}(Q,K,V)=\text{softmax}\left(\frac{QK^T}{\sqrt{d_k}}\right)V,
\end{aligned}
\label{eq:attention}
\end{equation}
where $Q$, $K$, and $V$ denote the query, key, and value matrices, respectively, with $d_k$ representing the dimensionality of the key vectors. 
In self-attention, these matrices are all derived from the same input feature map, enabling intra-modality feature integration. 
Conversely, in cross-attention, $Q$ is typically projected from the visual feature space, while $K$ and $V$ originate from the conditioning context (e.g., text embeddings), facilitating inter-modality information fusion.

\subsection{Global Attention Regulation}
\label{sec3.2:global attention regulation}


As substantiated by prior research~\cite{prompt,survey}, self-attention projections within Latent Diffusion Model (LDM)-based generative frameworks~\cite{sd2,sd3} inherently encapsulate both spatial positional relationships and semantic representations pertinent to content-related information. 
Consequently, the preservation of content image identity information during style transfer operations emerges as a critical prerequisite for maintaining structural fidelity. 
To address this fundamental requirement, we propose a novel feature fusion module that strategically leverages complementary information from both content teacher models and style teacher models, as shown in Fig.~\ref{fig:main}. 
This module enables global modulation of the self-attention layers within the main branch's stylized generation model, thereby ensuring dual objectives: 
1) consistent retention of content identity characteristics throughout the stylization process, and 
2) effective incorporation of stylistic attributes derived from either image or textual style references.
Specifically, the initial phase of our global attention regulation module focuses on synergistic integration of multi-source feature representations. 
Given the imperative for training-free feature manipulation, we employ adaptive instance normalization~\cite{Huang_2017_ICCV}, a pioneering work in the style transfer domain, to achieve decoupled feature recombination. 
This operation fuses content-specific attention projections $(Q_{self}^c, K_{self}^c, V_{self}^c)$ from content teacher models with style-specific projections $(Q_{self}^s, K_{self}^s, V_{self}^s)$ from style teacher models, generating optimized composite projections $(Q_{self}^{cs}, K_{self}^{cs}, V_{self}^{cs})$. 
The mathematical formalization of this fusion process, which aligns feature distribution statistics while preserving discriminative attributes, is detailed in Eq.~\ref{eq:3.2.1adain}.

{\small
\begin{equation}
\begin{aligned}
Q_{self}^{cs} &= \sigma\left(Q_{self}^s\right) \cdot \frac{Q_{self}^c - \mu\left(Q_{self}^c\right)}{\sigma\left(Q_{self}^c\right)} + \mu\left(Q_{self}^s\right), \\
K_{self}^{cs} &= \sigma\left(K_{self}^s\right) \cdot \frac{K_{self}^c - \mu\left(K_{self}^c\right)}{\sigma\left(K_{self}^c\right)} + \mu\left(K_{self}^s\right), \\
V_{self}^{cs} &= \sigma\left(V_{self}^s\right) \cdot \frac{V_{self}^c - \mu\left(V_{self}^c\right)}{\sigma\left(V_{self}^c\right)} + \mu\left(V_{self}^s\right),
\end{aligned}
\label{eq:3.2.1adain}
\end{equation}
}
where $Q_{self}^c, K_{self}^c, V_{self}^c$ are the content-specific attention projections in the content teacher model, $Q_{self}^s, K_{self}^s, V_{self}^s$ are the style-specific attention projections in the style teacher model, and $\mu\left(\cdot\right)$ and $\sigma\left(\cdot\right)$ are the mean and variance.


Building upon this foundational alignment, our methodology ensures that the statistical properties of the optimized composite projections $(Q_{self}^{cs}, K_{self}^{cs}, V_{self}^{cs})$ exhibit distributional congruence with the intrinsic self-attention projections $(Q_{self}^m, K_{self}^m, V_{self}^m)$ of the main branch. 
This critical correspondence establishes the theoretical basis for effective global modulation of the self-attention within the stylized generation model, enabling coordinated feature transformation throughout the denoising trajectory.
To implement this modulation, we employ a weighted fusion strategy that integrates the optimized composite projections with the main branch's native self-attention representations using a predefined hyperparameter. 
This controlled combination achieves dual objectives: 
1) persistent conservation of content identity information throughout stylization, and 
2) regulated incorporation of stylistic attributes from style references (style images or textual descriptions). 
The mathematical formulation in Eq.~\ref{eq:3.2.2reg} utilizes a fixed blending coefficient $\alpha$ to explicitly balance the trade-off between content preservation and style infusion, ensuring deterministic transformations independent of optimization dynamics.

    \begin{equation}
\begin{aligned}
\hat{Q}_{self}^{m} &= \alpha \cdot Q_{self}^m + (1 - \alpha) \cdot Q_{self}^{cs}, \\
\hat{K}_{self}^{m} &= \alpha \cdot K_{self}^m + (1 - \alpha) \cdot K_{self}^{cs}, \\
\hat{V}_{self}^{m} &= \alpha \cdot V_{self}^m + (1 - \alpha) \cdot V_{self}^{cs}, \\
\end{aligned}
\label{eq:3.2.2reg}
\end{equation}
where $\alpha$ is a hyperparameter used to control the weight of the fused attention projections.


Subsequently, the resulting modulated self-attention projections $(\hat{Q}_{self}^{m}, \hat{K}_{self}^{m}, \hat{V}_{self}^{m})$ are integrated into the main branch's stylized generation model's self-attention, ensuring that throughout the stylized generation process, the self-attention projections maintain content image identity information while simultaneously incorporating style references from style images/text.
The final self-attention expression is shown in Eq.~\ref{eq:3.2.3self-attention}.

\begin{equation}
\begin{aligned}
\text{Attention}(\hat{Q}_{self}^{m}, \hat{K}_{self}^{m}, \hat{V}_{self}^{m}), \\
\end{aligned}
\label{eq:3.2.3self-attention}
\end{equation}

\begin{figure*}[t]
\centering
\includegraphics[width=1\textwidth]{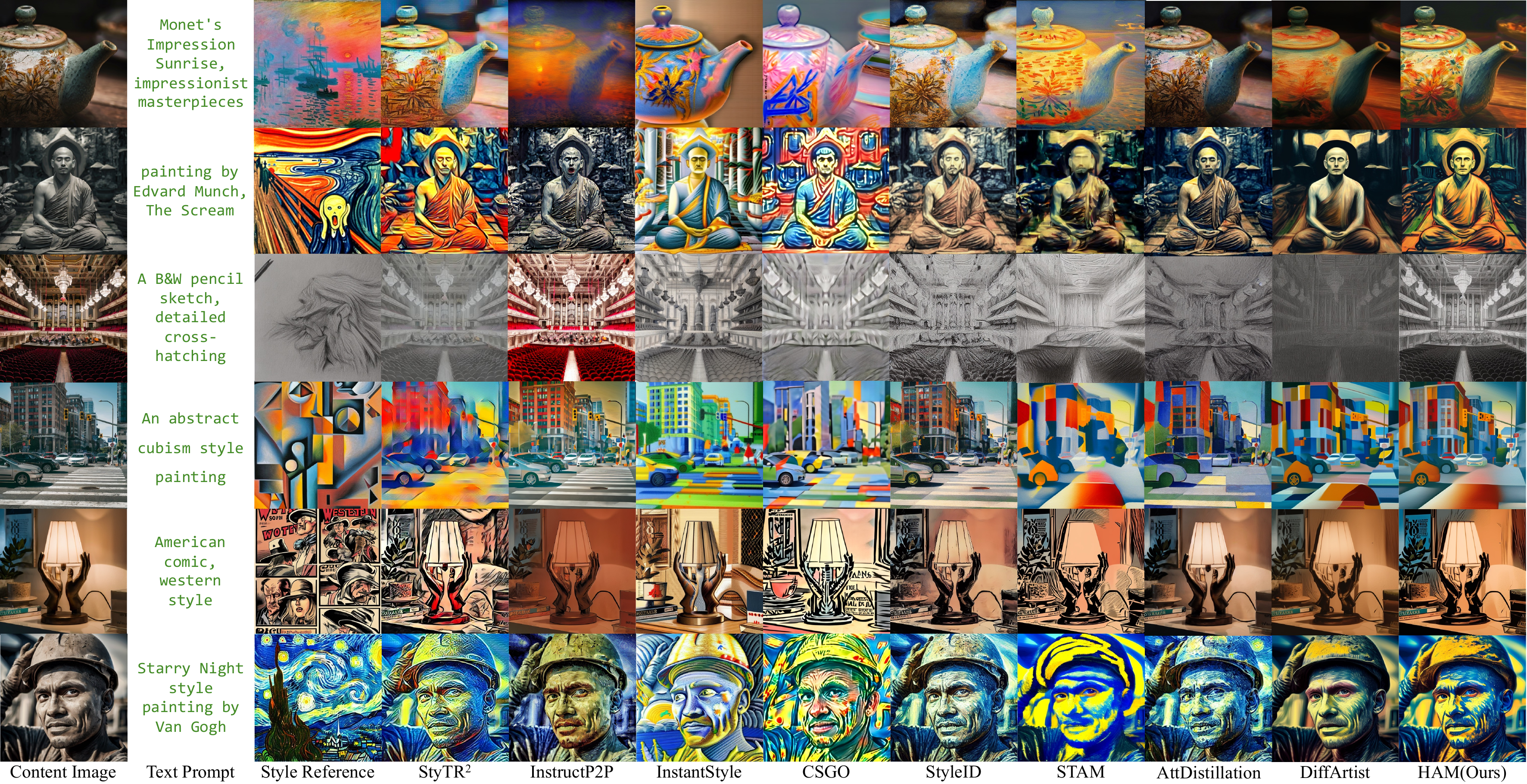}
\caption{Qualitative comparison with existing text-driven and image-driven SOTA methods. For fair evaluation, all methods use fixed random seeds: text-driven methods apply prompts directly, while image-driven methods generate style references via SD2.1 using identical prompts. Our HAM method better preserves content identity while maintaining style transfer semantics. 
}
\label{fig:main_exp}
\end{figure*}

\subsection{Local Attention Transplantation}
\label{sec3.3:local attention transplantation}


Existing methods for attention injection~\cite{styleid,stam} predominantly operate within the self-attention blocks of Stable Diffusion. 
However, since self-attention projections inherently encode substantial spatial-semantic structures, replacing key and value matrices in these blocks inevitably compromises content identity preservation. 
Furthermore, modifications to self-attention necessitate additional distribution alignment mechanisms (e.g., attention temperature scaling~\cite{styleid}) to mitigate discrepancies between query representations from different models and their corresponding key/value pairs.

To circumvent these limitations, we propose a novel paradigm shift: leveraging underutilized cross-attention channels for style transplantation. 
As illustrated in Fig.~\ref{fig:main}, our local attention transplantation module strategically employs feature representations extracted from both content and style teacher models. 
While prior methods~\cite{prompt,styleid} manipulate attention maps by substituting key/value projections in either cross-attention or self-attention blocks, our method also targets the cross-attention like them.
Concretely, we transplant style-specific key and value projections $(K_{cross}^s, V_{cross}^s)$ derived from style teacher models into the main stylization branch, replacing their native counterparts $(K_{cross}^m, V_{cross}^m)$ to achieve localized style injection.

Simultaneously, to prevent content identity degradation during diffusion and counteract potential style intrusion from transplanted projections, we implement a content protection mechanism for query representations. 
This is achieved through weighted fusion between content teacher model's query projections $Q_{cross}^c$ and the main branch's native query projections $Q_{cross}^m$, ensuring persistent conservation of structural identity. 
The complete operational formalization is provided in Eq.~\ref{eq:3.3cross-attention}.

\begin{equation}
\begin{aligned}
\hat{Q}_{cross}^m = \beta \cdot Q_{cross}^m + (1 - \beta) \cdot Q_{cross}^c, \\
\text{Attention}(\hat{Q}_{cross}^m, K_{cross}^s, V_{cross}^s), \\
\end{aligned}
\label{eq:3.3cross-attention}
\end{equation}
where $\beta$ is a hyperparameter that controls the query projection injection weight of the content teacher model.

\begin{table*}[t]
\centering
\resizebox{\textwidth}{!}{
\begin{tabular}{cccccccccc}
\hline
Method                                  & ArtFID$\downarrow$                                  & FID$\downarrow$                                    & LPIPS$\downarrow$                                  & LPIPS-Gray$\downarrow$                             & DINO$\uparrow$                                   & CLIP-I$\uparrow$                                 & CLIP-T$\uparrow$                                 & DC$\uparrow$                                     & CC$\uparrow$                                     \\ \hline
DDIM(ICLR'21)                     & \cellcolor[HTML]{FFFFFF}31.149          & \cellcolor[HTML]{FFFFFF}17.939         & \cellcolor[HTML]{D5EAF1}0.645          & \cellcolor[HTML]{FCFDFE}0.554          & \cellcolor[HTML]{FFFFFF}0.278          & \cellcolor[HTML]{FCFEFE}0.493          & \cellcolor[HTML]{FFFFFF}0.192          & \cellcolor[HTML]{FFFFFF}1.524          & \cellcolor[HTML]{FFFFFF}1.780          \\
ControlNet(ICCV'23)               & \cellcolor[HTML]{C4E1EB}24.751          & \cellcolor[HTML]{BBDCE8}13.472         & \cellcolor[HTML]{FFFFFF}0.710          & \cellcolor[HTML]{FFFFFF}0.557          & \cellcolor[HTML]{B3D9E6}0.513          & \cellcolor[HTML]{B7DBE8}0.583          & \cellcolor[HTML]{A9D4E3}0.210          & \cellcolor[HTML]{B3D9E7}1.831          & \cellcolor[HTML]{B7DBE8}1.916          \\
StyTR\textsuperscript{2}(CVPR'22) & \cellcolor[HTML]{81BFD5}17.460          & \cellcolor[HTML]{8CC5D9}10.433         & \cellcolor[HTML]{8AC4D8}0.527          & \cellcolor[HTML]{94C9DC}0.416          & \cellcolor[HTML]{CDE6EF}0.433          & \cellcolor[HTML]{FFFFFF}0.487          & \cellcolor[HTML]{BDDEEA}0.206          & \cellcolor[HTML]{CCE6EF}1.729          & \cellcolor[HTML]{F8FCFD}1.794          \\
InstructPix2Pix(CVPR'22)          & \cellcolor[HTML]{E4F1F6}28.319          & \cellcolor[HTML]{FAFCFD}17.657         & \cellcolor[HTML]{84C1D7}0.518          & \cellcolor[HTML]{94C9DC}0.415          & \cellcolor[HTML]{9ECFE0}0.575          & \cellcolor[HTML]{9BCDDF}0.620          & \cellcolor[HTML]{A4D1E1}0.211          & \cellcolor[HTML]{A0CFE0}1.908          & \cellcolor[HTML]{9ECFE0}1.963          \\
InstantStyle(arxiv'24)            & \cellcolor[HTML]{DBECF3}27.244          & \cellcolor[HTML]{D6EAF1}15.249         & \cellcolor[HTML]{E9F4F8}0.677          & \cellcolor[HTML]{FDFEFE}0.556          & \cellcolor[HTML]{C0DFEB}0.474          & \cellcolor[HTML]{A7D3E3}0.604          & \cellcolor[HTML]{E5F2F7}0.198          & \cellcolor[HTML]{C3E1EC}1.765          & \cellcolor[HTML]{B5DAE7}1.921          \\
CSGO(arxiv'24)                    & \cellcolor[HTML]{D9ECF2}27.116          & \cellcolor[HTML]{D5EAF1}15.207         & \cellcolor[HTML]{E7F3F7}0.673          & \cellcolor[HTML]{E7F3F7}0.527          & \cellcolor[HTML]{BDDEEA}0.482          & \cellcolor[HTML]{B9DCE8}0.581          & \cellcolor[HTML]{E7F3F7}0.197          & \cellcolor[HTML]{C1E0EB}1.775          & \cellcolor[HTML]{C3E1EC}1.893          \\
StyleID(CVPR'24)                  & \cellcolor[HTML]{6CB5CF}\underline{15.161}    & \cellcolor[HTML]{6CB5CF}\textbf{8.273} & \cellcolor[HTML]{CFE6EF}0.635          & \cellcolor[HTML]{E0EFF4}0.516          & \cellcolor[HTML]{A9D4E3}0.544          & \cellcolor[HTML]{9CCDDF}0.619          & \cellcolor[HTML]{9CCEDF}0.213          & \cellcolor[HTML]{A8D4E3}1.873          & \cellcolor[HTML]{9ECEE0}1.964          \\
STAM(CVPR'25)                     & \cellcolor[HTML]{7CBDD4}16.941          & \cellcolor[HTML]{7BBCD3}9.269          & \cellcolor[HTML]{D8EBF2}0.650          & \cellcolor[HTML]{ECF5F8}0.532          & \cellcolor[HTML]{ADD6E5}0.531          & \cellcolor[HTML]{A4D2E2}0.608          & \cellcolor[HTML]{77BBD3}\underline{0.221}    & \cellcolor[HTML]{A9D4E3}1.869          & \cellcolor[HTML]{9ECFE0}1.963          \\
AttDistillation(CVPR'25)          & \cellcolor[HTML]{75B9D2}16.170          & \cellcolor[HTML]{75BAD2}\underline{8.926}    & \cellcolor[HTML]{CBE5EE}0.629          & \cellcolor[HTML]{DEEEF4}0.514          & \cellcolor[HTML]{AAD4E3}0.541          & \cellcolor[HTML]{9FCFE0}0.615          & \cellcolor[HTML]{80BFD6}0.219          & \cellcolor[HTML]{A7D3E3}1.878          & \cellcolor[HTML]{9BCDDF}1.969          \\
DiffArtist(MM'25)                 & \cellcolor[HTML]{75B9D2}16.174          & \cellcolor[HTML]{80BFD5}9.641          & \cellcolor[HTML]{86C2D7}\underline{0.520}    & \cellcolor[HTML]{92C8DB}\underline{0.413}    & \cellcolor[HTML]{8DC6DA}\underline{0.629}    & \cellcolor[HTML]{97CBDD}\underline{0.626}    & \cellcolor[HTML]{7BBDD4}0.220          & \cellcolor[HTML]{8CC5DA}\underline{1.987}    & \cellcolor[HTML]{93C9DC}\underline{1.984}    \\
HAM(Ours)                         & \cellcolor[HTML]{6CB5CF}\textbf{15.151} & \cellcolor[HTML]{7ABCD3}9.244          & \cellcolor[HTML]{6CB5CF}\textbf{0.479} & \cellcolor[HTML]{6CB5CF}\textbf{0.362} & \cellcolor[HTML]{6CB5CF}\textbf{0.728} & \cellcolor[HTML]{6CB5CF}\textbf{0.682} & \cellcolor[HTML]{6CB5CF}\textbf{0.223} & \cellcolor[HTML]{6CB5CF}\textbf{2.113} & \cellcolor[HTML]{6CB5CF}\textbf{2.057} \\ \hline
\end{tabular}
}
\caption{Quantitative comparison with existing text-driven and image-driven SOTA methods. The best results are highlighted in bold, and the second best results are underlined. The color coding indicates relative performance.}
\label{tab:main_tab}
\end{table*}

\subsection{Style-Infused Noise Initialization}
\label{sec3.4:style-infused noise initialization}



As comprehensively delineated in Fig.~\ref{fig:main}, our framework achieves robust style transfer control throughout the diffusion process via the synergistic operation of global attention regulation and local attention transplantation modules. 
The configuration of initial noise consequently emerges as the pivotal remaining determinant of stylization efficacy, given its fundamental role in establishing the generative trajectory's starting state. 
While direct transplantation of the content teacher model's initial noise $z_0^c$ to the main stylization branch represents a conceptually straightforward approach, our quantitative evaluations and qualitative assessments (observed under $\gamma=1$ settings in Fig.~\ref{fig:ablation_hyper} and Tab.~\ref{tab:ablation_gamma}) demonstrate its inability to achieve meaningful style transfer, primarily attributable to insufficient style integration.
Similarly, adaptive instance normalization-based fusion of content noise $z_0^c$ and style noise $z_0^s$ yields composite initial noise $z_0^m$, yet empirical analysis reveals pronounced content identity degradation (observed under $\gamma=0$ settings in Fig.~\ref{fig:ablation_hyper} and Tab.~\ref{tab:ablation_gamma}). 
This phenomenon indicates an inherent optimization conflict between style intensity and content fidelity in conventional fusion paradigms.

To resolve this fundamental limitation, we innovate style-infused noise initialization that incorporates a dedicated content residual noise component atop the baseline AdaIN-fused stylized noise. 
This dual-component architecture explicitly balances stylization intensity against content preservation, enabling precise calibration of their relative contributions throughout the denoising cascade. 
The complete mathematical implementation is formalized in Eq.~\ref{eq:3.4adain}.

\begin{equation}
\begin{aligned}
z_T^m = \gamma \cdot &\underbrace{\left[ z_T^c - \left( \sigma(z_T^s) \cdot \frac{z_T^c - \mu(z_T^c)}{\sigma(z_T^c)} + \mu(z_T^s) \right) \right]}_{\text{Content Residual Noise}} \\+ &\underbrace{\left[ \sigma(z_T^s) \cdot \frac{z_T^c - \mu(z_T^c)}{\sigma(z_T^c)} + \mu(z_T^s) \right]}_{\text{Stylized Initial Noise}}, \\
\end{aligned}
\label{eq:3.4adain}
\end{equation}
where $\gamma$ is a hyper-parameter that controls the weight of the content residual noise, $z_T^c$ is the initial noise of the content teacher model, $z_T^s$ is the initial noise of the style teacher model, $z_T^m$ is the stylized initial noise for the main branch, and $\mu\left(\cdot\right)$ and $\sigma\left(\cdot\right)$ are the mean and variance.

\section{Experiments}
\label{sec4:experiments}

\subsection{Experiment Setup}
\label{sec4.1:experiment setup}


\paragraph{Implementation Details} 
We employ diffusion models based on the SD2.1 and SD3.5 architectures. 
The denoising process utilizes 50 steps for SD2.1. 
Images are resized to $512\times512$ pixels, with SD2.1 hyperparameters configured as $\alpha=0.75$, $\beta=0.25$, $\gamma=0.5$. 
Experiments execute on a single NVIDIA RTX3090, where SD2.1 inversion requires 4s (50 steps) and stylized image generation completes in 16s (50 steps).
The results of our method HAM on SD3.5 are discussed in the supplementary material.

\paragraph{Datasets} 
We conduct experiments on the MS-COCO~\cite{lin2014microsoft} and the WikiArt~\cite{artgan2018}. 
Specifically, 1,000 images from MS-COCO were randomly selected as the test content images. 
For WikiArt, a collection of images from multiple artists was chosen as the style references to represent the style distribution for FID computation.
The final test dataset comprises 1,000 content images, 1,000 corresponding generated stylized images, and the corresponding artists' works from WikiArt as style image references.


\paragraph{Comparison Methods} 
We compare with existing text-driven and image-driven methods: DDIM~\cite{ddim}, ControlNet~\cite{controlnet}, StyTR\textsuperscript{2}~\cite{deng2022stytr2}, InstructPix2Pix~\cite{brooks2023instructpix2pix}, InstantStyle~\cite{wang2024instantstyle}, CSGO~\cite{xing2024csgo}, StyleID~\cite{styleid}, STAM~\cite{stam}, AttDistillation~\cite{AttentionDistillation_2025_CVPR} and DiffArtist~\cite{artist}. 
For fair text/image-guided comparison, we fix random seeds and use identical prompts: 
(1) Text-guided: direct prompt input 
(2) Image-guided: prompts generate style references via SD2.1

\paragraph{Evaluation Metrics} 
We evaluate both traditional metrics FID~\cite{heusel2017gans}, LPIPS~\cite{zhang2018unreasonable}, ArtFID~\cite{wright2022artfid} and metrics based on DINO~\cite{oquab2023dinov2} and CLIP~\cite{radford2021learning} (DINO, CLIP-I, CLIP-T). 
For CLIP-T, the input text prompts are style-specific prompts. 
Our evaluation framework comprises: 
\textbf{(1) Style strength metrics}: \textbf{FID} and \textbf{CLIP-T} measuring stylization degree;
\textbf{(2) Content preservation metrics}: \textbf{LPIPS}, \textbf{DINO} and \textbf{CLIP-I} assessing content consistency;
\textbf{(3) Comprehensive style transfer metric}: \textbf{ArtFID} for overall performance. 
To better comprehensively evaluate stylized images, we introduce two novel composite metrics following ArtFID's computation paradigm:
$\textbf{DC}=(\text{DINO}+1)\cdot(\text{CLIP-T}+1)$ and $\textbf{CC}=(\text{CLIP-I}+1)\cdot(\text{CLIP-T}+1)$.




\begin{figure}[t]
\centering
\includegraphics[width=1\columnwidth]{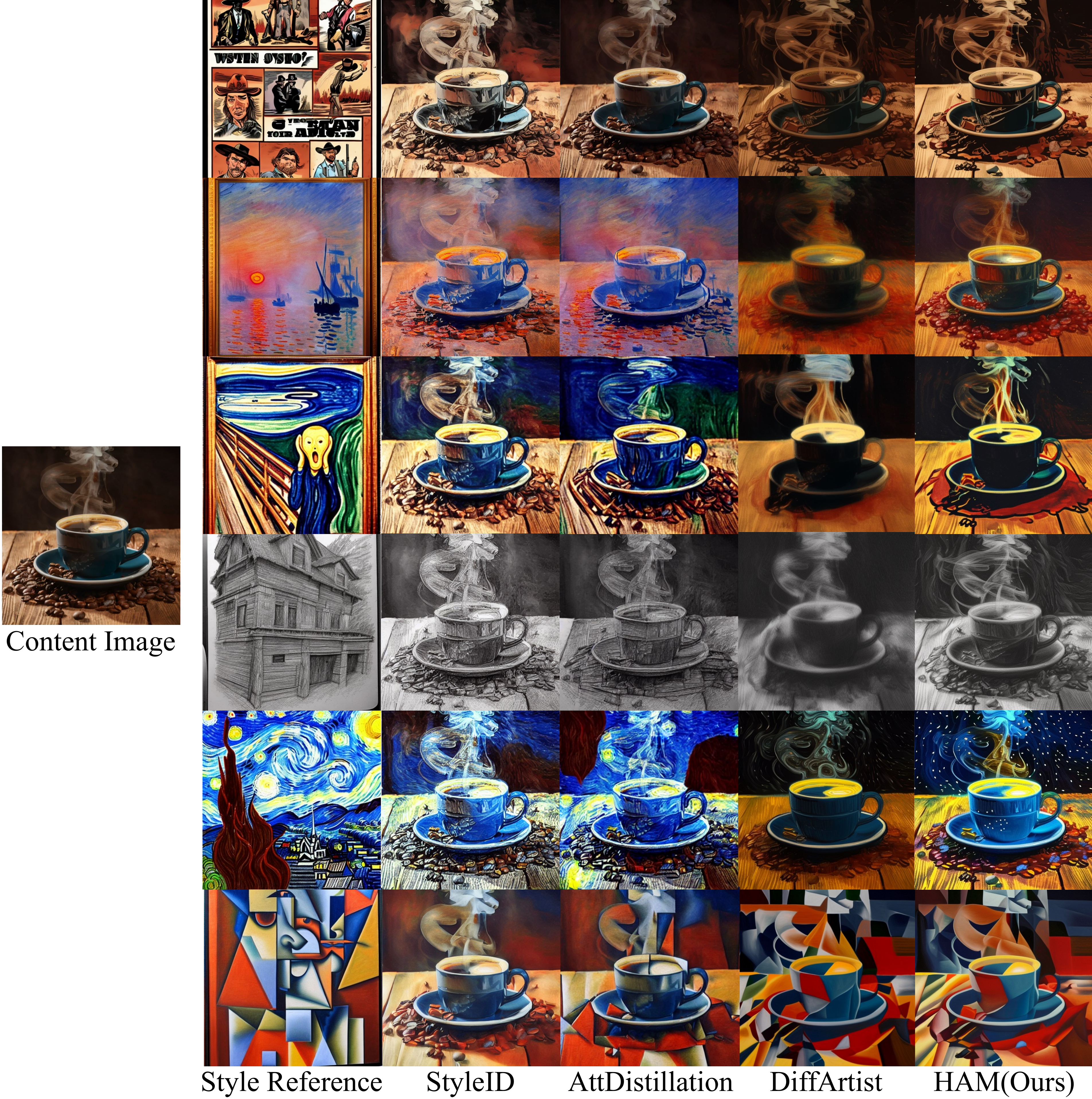}
\caption{Qualitative results of our method HAM and the SOTA method are presented under different style references for the same content image. It can be observed that our method HAM has significant advantages in both style transfer and identity preservation.
}
\label{fig:main_show}
\end{figure}

\subsection{Performance Evaluation}
\label{sec4.2:performance evaluation}

\paragraph{Qualitative Evaluation}

As shown in Fig.~\ref{fig:main_exp}, under our fair evaluation setting, our HAM and existing SOTA methods are compared across diverse text-driven and image-driven style transfer tasks. 
Due to space constraints, qualitative results omit several methods (e.g., DDIM, ControlNet) that perform poorly in quantitative assessments.
Visual results indicate that StyTR\textsuperscript{2} and InstructPix2Pix struggle to capture complex style patterns, leading to noticeable style leakage and content distortion. 
InstantStyle and CSGO either inadequately represent style details or suffer from severe style-content leakage, resulting in loss of identity. 
Although StyleID, STAM, AttDistillation, DiffArtist produce reasonable stylizations, they exhibit a consistent style-content trade-off: either style is well-captured at the cost of content structure, or content is preserved with insufficient style expression.
In contrast, HAM accurately captures stylistic attributes under various style-references while maintaining high content fidelity, thereby improving the overall quality of the generated stylized images.


Additionally, as illustrated in Fig.~\ref{fig:main_show}, another qualitative experiment is conducted using a single content image under varied style references, with fixed parameters across all methods and no specific adjustments for any styles.
The results reveal that existing SOTA methods not only struggle with balancing style and content but also demonstrate limited adaptability when applied to diverse styles using the same content image.
In contrast, our HAM shows stronger robustness in transferring complex stylistic information from multiple references to the same content image, while more effectively preserving structural details compared to current SOTA methods.
These findings affirm the robustness of our HAM in handling style transfer tasks.

\paragraph{Quantitative Evaluation}
As shown in Tab.~\ref{tab:main_tab}, our method HAM achieves optimal (CLIP-T) and near-optimal (FID) performance on two critical style-strength metrics, demonstrating its efficacy in distilling and transferring style reference information.
Crucially, the top-ranked CLIP-T performance explicitly confirms HAM's exceptional alignment with textual style semantics across diverse prompts. 
For traditional content preservation metrics (LPIPS and LPIPS-Gray), HAM significantly outperforms all baselines by notable margins, underscoring its absolute advantage in retaining both structural integrity and fine-grained details of content images before and after color removal. 
This quantitatively reinforces qualitative observations of high content identity preservation. 
Furthermore, the other content metrics (DINO and CLIP-I) validate HAM's robustness in preserving inter-image structural coherence, global visual consistency, and visual semantic consistency. 
This is consistent with our qualitative conclusion regarding generation performance.
Finally, regarding overall quality metrics (ArtFID, DC, CC), HAM maintains optimal balance between holistic content preservation and precise style semantics, markedly surpassing all competing methods in synthesizing perceptually harmonious outputs.

\begin{figure}[t]
\centering
\includegraphics[width=1\columnwidth]{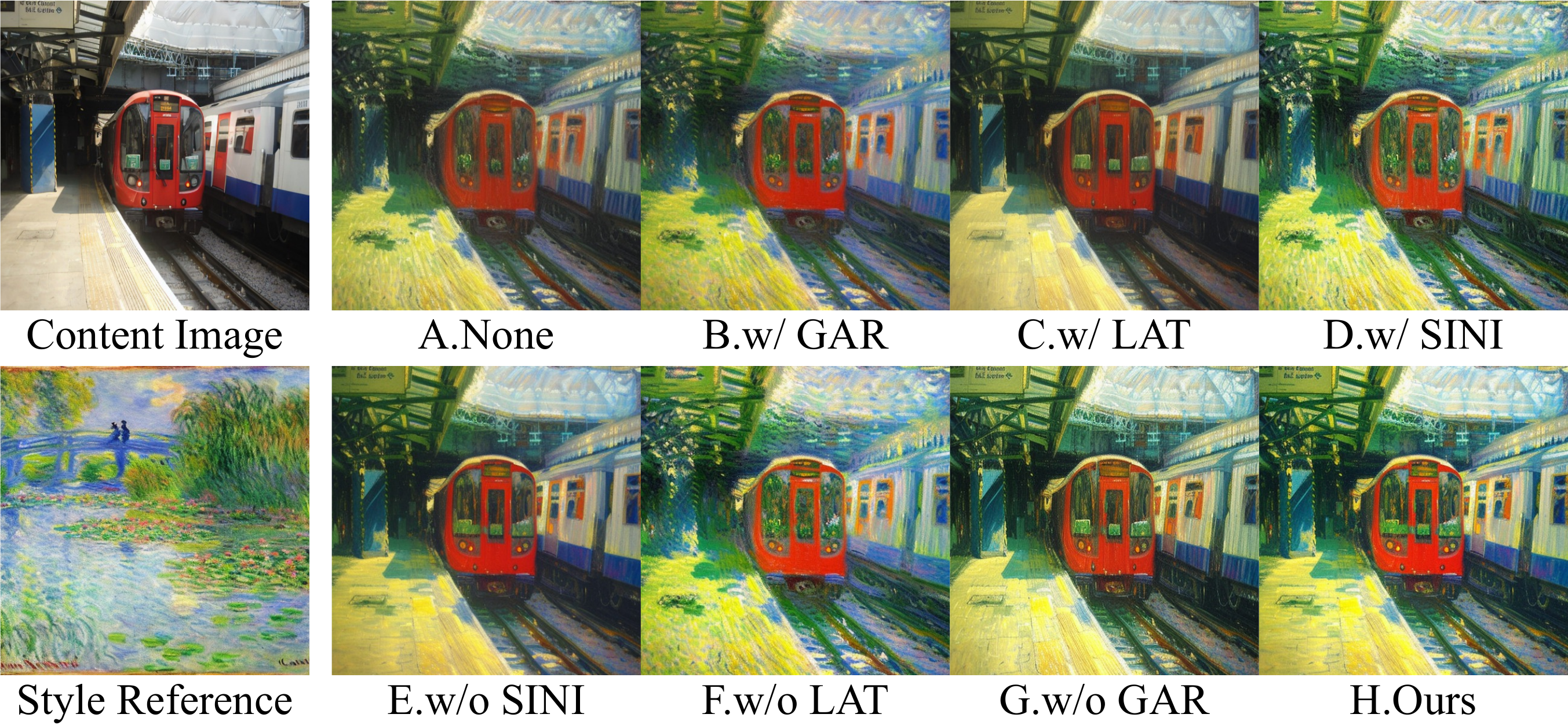}
\caption{Qualitative ablation study of different modules in our method. The indexes are consistent with those in the quantitative experiments.}
\label{fig:ablation}
\end{figure}
\begin{figure}[t]
\centering
\includegraphics[width=1\columnwidth]{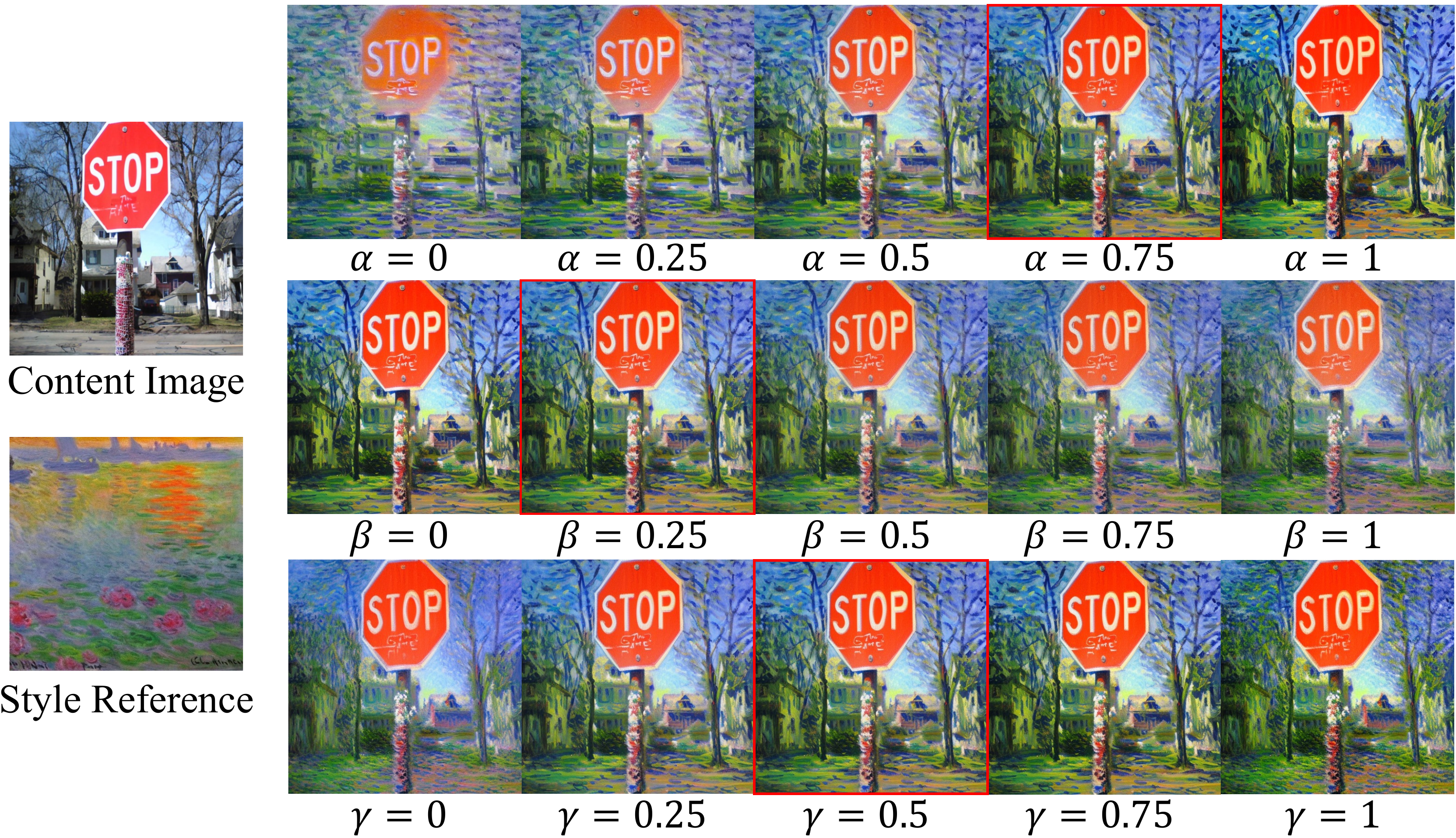}
\caption{Qualitative ablation study of hyper-parameters in our method. The red boxes represent the hyperparameters we selected.
}
\label{fig:ablation_hyper}
\end{figure}

\begin{table}[t]
\centering
\resizebox{\columnwidth}{!}{
\begin{tabular}{c|ccc|ccccc}
\hline
No. & GAR & LAT & SINI & DINO                                   & CLIP-I                                 & CLIP-T                                 & DC                                     & CC                                     \\ \hline
A   & \ding{55}  & \ding{55}  & \ding{55}   & 0.609          & 0.626          & 0.220          & 1.963          & 1.984          \\
B   & \ding{51} & \ding{55}  & \ding{55}   & 0.618          & 0.626          & \underline{0.231}    & 1.993          & 2.002          \\
C   & \ding{55}  & \ding{51} & \ding{55}   & 0.712          & \underline{0.696}    & 0.193          & 2.042          & 2.023          \\
D   & \ding{55}  & \ding{55}  & \ding{51}  & 0.588          & 0.624          & 0.225          & 1.945          & 1.989          \\
E   & \ding{51} & \ding{51} & \ding{55}   & \textbf{0.746} & \textbf{0.696} & 0.202          & \underline{2.099}    & \underline{2.040}    \\
F   & \ding{51} & \ding{55}  & \ding{51}  & 0.599          & 0.627          & \textbf{0.235} & 1.976          & 2.010          \\
G   & \ding{55}  & \ding{51} & \ding{51}  & 0.695          & 0.679          & 0.200          & 2.034          & 2.015          \\
H   & \ding{51} & \ding{51} & \ding{51}  & \underline{0.728}    & 0.682          & 0.223          & \textbf{2.113} & \textbf{2.057} \\ \hline
\end{tabular}
}
\caption{Quantitative ablation study of different modules in our method. The indexes are the same as those in the qualitative experiments.}
\label{tab:ablation_tab}
\end{table}
\begin{table}[t]
\centering
\resizebox{\columnwidth}{!}{
\begin{tabular}{cccccc}
\hline
$\alpha$    & DINO$\uparrow$                          & CLIP-I$\uparrow$                        & CLIP-T$\uparrow$                        & DC$\uparrow$                          & CC$\uparrow$                          \\ \hline
1                               & \underline{0.695} & \underline{0.679} & 0.200 & 2.034 & 2.015 \\
0.75                            & \textbf{0.728} & \textbf{0.682} & 0.223 & \textbf{2.113} & \textbf{2.057} \\
0.5                             & 0.690 & 0.667 & 0.225 & \underline{2.070} & \underline{2.042} \\
0.25                            & 0.602 & 0.634 & \underline{0.226} & 1.964 & 2.003 \\
0                               & 0.497 & 0.602 & \textbf{0.229} & 1.840 & 1.969 \\ \hline
\end{tabular}
}
\caption{Quantitative ablation study of hyper-parameter $\alpha$ in Global Attention Regulation (GAR).}
\label{tab:ablation_alpha}
\vspace{-1em}
\end{table}

\subsection{Ablation Study}
\label{sec4.4:ablation study}

\paragraph{Ablation on different modules}
As illustrated in Fig.~\ref{fig:ablation} and Tab.~\ref{tab:ablation_tab}, ablation studies evaluate three core modules of the HAM method.
Both qualitative and quantitative results demonstrate the contributions of our modules. 
Global attention regulation module improves the CLIP-T score while slightly increasing DINO and CLIP-I metrics. 
Qualitative results further illustrate its role in enhancing the stylization intensity and content preservation, confirming its consistency with our design objective.
This is also consistent with the performance of quantitative indicators.
The local attention transplantation module significantly improves DINO and CLIP-I content preservation metrics but adversely affects CLIP-T. 
This corresponds to qualitatively observed stronger identity retention and reduced stylization, indicating that integrating the style teacher's key/value requires protecting the main branch's query with the content teacher's query features (further evidenced in the next paragraph for the hyperparameter $\beta$).
For style-infused noise initialization, a modest CLIP-T improvement is observed with no measurable impact on DINO or CLIP-I. 
Visually, this module enriches stylistic elements and color diversity in the output.
Similarly, the results of quantitative and qualitative experiments on SINI are consistent.
Collectively, the modules establish HAM's balance on DC/CC metrics.

\paragraph{Ablation on different hyper-parameters}
As shown in Fig.~\ref{fig:ablation_hyper}, ablation experiments assess HAM's key hyperparameters $(\alpha,\beta,\gamma)$ qualitatively.  
For $\alpha$ in the global attention regulation module, as shown in Tab.~\ref{tab:ablation_alpha},content preservation metrics (DINO and CLIP-I) achieve optimal performance at $\alpha=0.75$, degrading at other values, consistent with qualitative observations notably regarding text below STOP signs. 
Conversely, CLIP-T increases as $\alpha$ decreases, and qualitatively, the generated image tends to be stylized, consistent with qualitative assessments and design intent.
DC and CC confirm $\alpha=0.75$ yields best overall performance.  
In the local attention transfer module, as shown in Tab.~\ref{tab:ablation_beta},for parameter $\beta$, decreasing $\beta$ can improve the content score (DINO, CLIP-I) while decreasing CLIP-T. 
Qualitatively, this manifests in the generated image as text and other main subjects gradually becoming blurred, increasing the stylization of the image, consistent with the quantitative results.
$\beta$ weights the query injection, governing content protection versus style transfer, mutually constrained. 
Based on DC and CC, $\beta=0.25$ provides optimal balance, validated quantitatively and qualitatively.  
For $\gamma$ in the style-infused noise initialization module, as shown in Tab.~\ref{tab:ablation_gamma}, CLIP-T peaks at $\gamma=0.5$, while content metrics decline as $\gamma$ decreases, consistent with enhanced style information at content identity expense. 
DC and CC indicate $\gamma=0.5$ offers best stylization outcome, validated quantitatively and qualitatively.

\begin{table}[t]
\centering
\resizebox{\columnwidth}{!}{
\begin{tabular}{cccccc}
\hline
$\beta$  & DINO$\uparrow$                          & CLIP-I$\uparrow$                        & CLIP-T$\uparrow$                        & DC$\uparrow$                          & CC$\uparrow$                          \\ \hline
1                               & 0.599 & 0.627 & \textbf{0.235} & 1.976 & 2.010 \\
0.75                            & 0.646 & 0.630 & \underline{0.227} & 2.019 & 1.999 \\
0.5                             & 0.689 & 0.650 & 0.226 & 2.071 & 2.023 \\
0.25                            & \underline{0.728} & \underline{0.682} & 0.223 & \textbf{2.113} & \textbf{2.057} \\
0                               & \textbf{0.739} & \textbf{0.704} & 0.201 & \underline{2.088} & \underline{2.046} \\ \hline
\end{tabular}
}
\caption{Quantitative ablation study of hyper-parameter $\beta$ in Local Attention Transplantation (LAT).}
\label{tab:ablation_beta}
\end{table}
\begin{table}[t]
\centering
\resizebox{\columnwidth}{!}{
\begin{tabular}{cccccc}
\hline
$\gamma$       & DINO$\uparrow$                         & CLIP-I$\uparrow$                        & CLIP-T$\uparrow$                        & DC$\uparrow$                          & CC$\uparrow$                          \\ \hline
1                               & \textbf{0.746} & \textbf{0.696} & 0.202 & 2.099 & 2.040 \\
0.75                            & \underline{0.733} & \underline{0.689} & 0.212 & \underline{2.101} & \underline{2.048} \\
0.5                             & 0.728 & 0.682 & \textbf{0.223} & \textbf{2.113} & \textbf{2.057} \\
0.25                            & 0.714 & 0.678 & \underline{0.217} & 2.086 & 2.042 \\
0                               & 0.708 & 0.674 & 0.212 & 2.070 & 2.029 \\ \hline
\end{tabular}
}
\caption{Quantitative ablation study of hyper-parameter $\gamma$ in Style-Infused Noise Initialization (SINI).}
\label{tab:ablation_gamma}
\vspace{-1em}
\end{table}

\section{Conclusion and Limitations}
\label{sec5:conclusion}

We propose HAM, a training-free style transfer framework that addresses the core challenge of content-style balance via heterogeneous attention modulation. 
Our method begins with SINI, followed by two heterogeneous modulation mechanisms: GAR and LAT.
Together, these components work synergistically to achieve superior content-style equilibrium. 
Extensive experiments demonstrate that HAM outperforms the state-of-the-art methods in both fidelity and stylistic quality. 
Although our method advances content-style balancing, transferring highly abstract or surrealistic artistic styles remains an open challenge for future work.

\section*{Acknowledgement}
This work was supported by the National Natural Science Foundation of China (62322211, 62336008), the “Pioneer” and “Leading Goose” R\&D Program of Zhejiang Province(2024C01023).

{
    \small
    \bibliographystyle{ieeenat_fullname}
    \bibliography{main}

@String(CVPR= {IEEE Conf. Comput. Vis. Pattern Recog.})

@String(ICCV= {Int. Conf. Comput. Vis.})

@String(ICLR = {Int. Conf. Learn. Represent.})

@String(AAAI = {AAAI})

@String(CVPR  = {CVPR})

@String(ICCV  = {ICCV})

@String(ICLR  = {ICLR})

@InProceedings{sd2,
    author    = {Rombach, Robin and Blattmann, Andreas and Lorenz, Dominik and Esser, Patrick and Ommer, Bj\"orn},
    title     = {High-Resolution Image Synthesis With Latent Diffusion Models},
    booktitle = {Proceedings of the IEEE/CVF Conference on Computer Vision and Pattern Recognition (CVPR)},
    month     = {June},
    year      = {2022},
    pages     = {10684-10695}
}

@inproceedings{sd3,
title={Scaling Rectified Flow Transformers for High-Resolution Image Synthesis},
author={Patrick Esser and Sumith Kulal and Andreas Blattmann and Rahim Entezari and Jonas M{\"u}ller and Harry Saini and Yam Levi and Dominik Lorenz and Axel Sauer and Frederic Boesel and Dustin Podell and Tim Dockhorn and Zion English and Robin Rombach},
booktitle={Forty-first International Conference on Machine Learning},
year={2024},
url={https://openreview.net/forum?id=FPnUhsQJ5B}
}

@article{sdxl,
  title={Sdxl: Improving latent diffusion models for high-resolution image synthesis},
  author={Podell, Dustin and English, Zion and Lacey, Kyle and Blattmann, Andreas and Dockhorn, Tim and M{\"u}ller, Jonas and Penna, Joe and Rombach, Robin},
  journal={arXiv preprint arXiv:2307.01952},
  year={2023}
}

@article{ddim,
  title={Denoising diffusion implicit models},
  author={Song, Jiaming and Meng, Chenlin and Ermon, Stefano},
  journal={arXiv preprint arXiv:2010.02502},
  year={2020}
}

@inproceedings{dalle,
  title={Zero-shot text-to-image generation},
  author={Ramesh, Aditya and Pavlov, Mikhail and Goh, Gabriel and Gray, Scott and Voss, Chelsea and Radford, Alec and Chen, Mark and Sutskever, Ilya},
  booktitle={International conference on machine learning},
  pages={8821--8831},
  year={2021},
  organization={Pmlr}
}

@article{dalle2,
  title={Hierarchical text-conditional image generation with clip latents},
  author={Ramesh, Aditya and Dhariwal, Prafulla and Nichol, Alex and Chu, Casey and Chen, Mark},
  journal={arXiv preprint arXiv:2204.06125},
  volume={1},
  number={2},
  pages={3},
  year={2022}
}

@article{dalle3,
  title={Improving image generation with better captions},
  author={Betker, James and Goh, Gabriel and Jing, Li and Brooks, Tim and Wang, Jianfeng and Li, Linjie and Ouyang, Long and Zhuang, Juntang and Lee, Joyce and Guo, Yufei and others},
  journal={Computer Science. https://cdn. openai. com/papers/dall-e-3. pdf},
  volume={2},
  number={3},
  pages={8},
  year={2023}
}

@inproceedings{styleid,
  title={Style injection in diffusion: A training-free approach for adapting large-scale diffusion models for style transfer},
  author={Chung, Jiwoo and Hyun, Sangeek and Heo, Jae-Pil},
  booktitle={Proceedings of the IEEE/CVF conference on computer vision and pattern recognition},
  pages={8795--8805},
  year={2024}
}

@inproceedings{artist,
author = {Jiang, Ruixiang and Chen, Chang Wen},
title = {DiffArtist: Towards Structure and Appearance Controllable Image Stylization},
year = {2025},
isbn = {9798400720352},
publisher = {Association for Computing Machinery},
address = {New York, NY, USA},
url = {https://doi.org/10.1145/3746027.3755010},
doi = {10.1145/3746027.3755010},
booktitle = {Proceedings of the 33rd ACM International Conference on Multimedia},
pages = {9598–9607},
numpages = {10},
location = {Dublin, Ireland},
series = {MM '25}
}

@InProceedings{Tumanyan_2023_CVPR,
    author    = {Tumanyan, Narek and Geyer, Michal and Bagon, Shai and Dekel, Tali},
    title     = {Plug-and-Play Diffusion Features for Text-Driven Image-to-Image Translation},
    booktitle = {Proceedings of the IEEE/CVF Conference on Computer Vision and Pattern Recognition (CVPR)},
    month     = {June},
    year      = {2023},
    pages     = {1921-1930}
}

@InProceedings{Kawar_2023_CVPR,
    author    = {Kawar, Bahjat and Zada, Shiran and Lang, Oran and Tov, Omer and Chang, Huiwen and Dekel, Tali and Mosseri, Inbar and Irani, Michal},
    title     = {Imagic: Text-Based Real Image Editing With Diffusion Models},
    booktitle = {Proceedings of the IEEE/CVF Conference on Computer Vision and Pattern Recognition (CVPR)},
    month     = {June},
    year      = {2023},
    pages     = {6007-6017}
}

@InProceedings{Liu_2024_CVPR,
    author    = {Liu, Chang and Li, Xiangtai and Ding, Henghui},
    title     = {Referring Image Editing: Object-level Image Editing via Referring Expressions},
    booktitle = {Proceedings of the IEEE/CVF Conference on Computer Vision and Pattern Recognition (CVPR)},
    month     = {June},
    year      = {2024},
    pages     = {13128-13138}
}

@InProceedings{Zhang_2023_CVPR,
    author    = {Zhang, Yuxin and Huang, Nisha and Tang, Fan and Huang, Haibin and Ma, Chongyang and Dong, Weiming and Xu, Changsheng},
    title     = {Inversion-Based Style Transfer With Diffusion Models},
    booktitle = {Proceedings of the IEEE/CVF Conference on Computer Vision and Pattern Recognition (CVPR)},
    month     = {June},
    year      = {2023},
    pages     = {10146-10156}
}

@InProceedings{Huang_2017_ICCV,
author = {Huang, Xun and Belongie, Serge},
title = {Arbitrary Style Transfer in Real-Time With Adaptive Instance Normalization},
booktitle = {Proceedings of the IEEE International Conference on Computer Vision (ICCV)},
month = {Oct},
year = {2017}
}

@misc{kwon2023diffusionbasedimagetranslationusing,
      title={Diffusion-based Image Translation using Disentangled Style and Content Representation}, 
      author={Gihyun Kwon and Jong Chul Ye},
      year={2023},
      eprint={2209.15264},
      archivePrefix={arXiv},
      primaryClass={cs.CV},
      url={https://arxiv.org/abs/2209.15264}, 
}

@inproceedings{frenkel2024implicit,
  title={Implicit style-content separation using b-lora},
  author={Frenkel, Yarden and Vinker, Yael and Shamir, Ariel and Cohen-Or, Daniel},
  booktitle={European Conference on Computer Vision},
  pages={181--198},
  year={2024},
  organization={Springer}
}

@article{lora,
  title={Lora: Low-rank adaptation of large language models.},
  author={Hu, Edward J and Shen, Yelong and Wallis, Phillip and Allen-Zhu, Zeyuan and Li, Yuanzhi and Wang, Shean and Wang, Lu and Chen, Weizhu and others},
  journal={ICLR},
  volume={1},
  number={2},
  pages={3},
  year={2022}
}

@inproceedings{controlnet,
  title={Adding conditional control to text-to-image diffusion models},
  author={Zhang, Lvmin and Rao, Anyi and Agrawala, Maneesh},
  booktitle={Proceedings of the IEEE/CVF international conference on computer vision},
  pages={3836--3847},
  year={2023}
}

@inproceedings{prompt,
  author={Amir Hertz and Ron Mokady and Jay Tenenbaum and Kfir Aberman and Yael Pritch and Daniel Cohen-Or},
  title={Prompt-to-Prompt Image Editing with Cross-Attention Control},
  year={2023},
  cdate={1672531200000},
  url={https://openreview.net/forum?id=_CDixzkzeyb},
  booktitle={ICLR},
}

@article{survey,
  title={A survey of multimodal-guided image editing with text-to-image diffusion models},
  author={Shuai, Xincheng and Ding, Henghui and Ma, Xingjun and Tu, Rongcheng and Jiang, Yu-Gang and Tao, Dacheng},
  journal={arXiv preprint arXiv:2406.14555},
  year={2024}
}

@inproceedings{lv2024pick,
  title={Pick-and-draw: Training-free semantic guidance for text-to-image personalization},
  author={Lv, Henglei and Xiao, Jiayu and Li, Liang},
  booktitle={Proceedings of the 32nd ACM International Conference on Multimedia},
  pages={10535--10543},
  year={2024}
}

@inproceedings{mo2024freecontrol,
  title={Freecontrol: Training-free spatial control of any text-to-image diffusion model with any condition},
  author={Mo, Sicheng and Mu, Fangzhou and Lin, Kuan Heng and Liu, Yanli and Guan, Bochen and Li, Yin and Zhou, Bolei},
  booktitle={Proceedings of the IEEE/CVF Conference on Computer Vision and Pattern Recognition},
  pages={7465--7475},
  year={2024}
}

@article{xing2024csgo,
  title={Csgo: Content-style composition in text-to-image generation},
  author={Xing, Peng and Wang, Haofan and Sun, Yanpeng and Wang, Qixun and Bai, Xu and Ai, Hao and Huang, Renyuan and Li, Zechao},
  journal={arXiv preprint arXiv:2408.16766},
  year={2024}
}

@misc{vae,
  title={Auto-encoding variational bayes},
  author={Kingma, Diederik P and Welling, Max and others},
  year={2013},
  publisher={Banff, Canada}
}

@article{attention,
  title={Attention is all you need},
  author={Vaswani, Ashish and Shazeer, Noam and Parmar, Niki and Uszkoreit, Jakob and Jones, Llion and Gomez, Aidan N and Kaiser, {\L}ukasz and Polosukhin, Illia},
  journal={Advances in neural information processing systems},
  volume={30},
  year={2017}
}

@inproceedings{stam,
  title={STAM: Zero-Shot Style Transfer using Diffusion Model via Attention Modulation},
  author={Fahim, Masud An Nur Islam and Saqib, Nazmus and Boutellier, Jani},
  booktitle={Proceedings of the Computer Vision and Pattern Recognition Conference},
  pages={6333--6343},
  year={2025}
}

@inproceedings{lin2014microsoft,
  title={Microsoft coco: Common objects in context},
  author={Lin, Tsung-Yi and Maire, Michael and Belongie, Serge and Hays, James and Perona, Pietro and Ramanan, Deva and Doll{\'a}r, Piotr and Zitnick, C Lawrence},
  booktitle={European conference on computer vision},
  pages={740--755},
  year={2014},
  organization={Springer}
}

@article{artgan2018,
  title={Improved ArtGAN for Conditional Synthesis of Natural Image and Artwork},
  author={Tan, Wei Ren and Chan, Chee Seng and Aguirre, Hernan and Tanaka, Kiyoshi},
  journal={IEEE Transactions on Image Processing},
  volume    = {28},
  number    = {1},
  pages     = {394--409},
  year      = {2019},
  url       = {https://doi.org/10.1109/TIP.2018.2866698},
  doi       = {10.1109/TIP.2018.2866698}
}

@article{heusel2017gans,
  title={Gans trained by a two time-scale update rule converge to a local nash equilibrium},
  author={Heusel, Martin and Ramsauer, Hubert and Unterthiner, Thomas and Nessler, Bernhard and Hochreiter, Sepp},
  journal={Advances in neural information processing systems},
  volume={30},
  year={2017}
}

@inproceedings{zhang2018unreasonable,
  title={The unreasonable effectiveness of deep features as a perceptual metric},
  author={Zhang, Richard and Isola, Phillip and Efros, Alexei A and Shechtman, Eli and Wang, Oliver},
  booktitle={Proceedings of the IEEE conference on computer vision and pattern recognition},
  pages={586--595},
  year={2018}
}

@inproceedings{wright2022artfid,
  title={Artfid: Quantitative evaluation of neural style transfer},
  author={Wright, Matthias and Ommer, Bj{\"o}rn},
  booktitle={DAGM German Conference on Pattern Recognition},
  pages={560--576},
  year={2022},
  organization={Springer}
}

@misc{oquab2023dinov2,
  title={DINOv2: Learning Robust Visual Features without Supervision},
  author={Oquab, Maxime and Darcet, Timothée and Moutakanni, Theo and Vo, Huy V. and Szafraniec, Marc and Khalidov, Vasil and Fernandez, Pierre and Haziza, Daniel and Massa, Francisco and El-Nouby, Alaaeldin and Howes, Russell and Huang, Po-Yao and Xu, Hu and Sharma, Vasu and Li, Shang-Wen and Galuba, Wojciech and Rabbat, Mike and Assran, Mido and Ballas, Nicolas and Synnaeve, Gabriel and Misra, Ishan and Jegou, Herve and Mairal, Julien and Labatut, Patrick and Joulin, Armand and Bojanowski, Piotr},
  journal={arXiv:2304.07193},
  year={2023}
}

@inproceedings{radford2021learning,
  title={Learning transferable visual models from natural language supervision},
  author={Radford, Alec and Kim, Jong Wook and Hallacy, Chris and Ramesh, Aditya and Goh, Gabriel and Agarwal, Sandhini and Sastry, Girish and Askell, Amanda and Mishkin, Pamela and Clark, Jack and others},
  booktitle={International conference on machine learning},
  pages={8748--8763},
  year={2021},
  organization={PmLR}
}

@inproceedings{deng2022stytr2,
  title={Stytr2: Image style transfer with transformers},
  author={Deng, Yingying and Tang, Fan and Dong, Weiming and Ma, Chongyang and Pan, Xingjia and Wang, Lei and Xu, Changsheng},
  booktitle={Proceedings of the IEEE/CVF conference on computer vision and pattern recognition},
  pages={11326--11336},
  year={2022}
}

@inproceedings{brooks2023instructpix2pix,
  title={Instructpix2pix: Learning to follow image editing instructions},
  author={Brooks, Tim and Holynski, Aleksander and Efros, Alexei A},
  booktitle={Proceedings of the IEEE/CVF conference on computer vision and pattern recognition},
  pages={18392--18402},
  year={2023}
}

@article{wang2024instantstyle,
  title={Instantstyle: Free lunch towards style-preserving in text-to-image generation},
  author={Wang, Haofan and Spinelli, Matteo and Wang, Qixun and Bai, Xu and Qin, Zekui and Chen, Anthony},
  journal={arXiv preprint arXiv:2404.02733},
  year={2024}
}

@InProceedings{AttentionDistillation_2025_CVPR,
    author    = {Zhou, Yang and Gao, Xu and Chen, Zichong and Huang, Hui},
    title     = {Attention Distillation: A Unified Approach to Visual Characteristics Transfer},
    booktitle = {Proceedings of the IEEE/CVF Conference on Computer Vision and Pattern Recognition (CVPR)},
    month     = {June},
    year      = {2025},
    pages     = {18270-18280}
}

@ARTICLE{11121533,
  author={Li, Liang and Cong, Gaoxiang and Qi, Yuankai and Zha, Zheng-Jun and Wu, Qi and Sheng, Quan Z. and Huang, Qingming and Yang, Ming-Hsuan},
  journal={IEEE Transactions on Pattern Analysis and Machine Intelligence}, 
  title={Dubbing Movies via Hierarchical Phoneme Modeling and Acoustic Diffusion Denoising}, 
  year={2025},
  volume={47},
  number={11},
  pages={10361-10377},
  doi={10.1109/TPAMI.2025.3597267}}

@ARTICLE{10607969,
  author={Zhang, Beichen and Li, Liang and Wang, Shuhui and Cai, Shaofei and Zha, Zheng-Jun and Tian, Qi and Huang, Qingming},
  journal={IEEE Transactions on Pattern Analysis and Machine Intelligence}, 
  title={Inductive State-Relabeling Adversarial Active Learning With Heuristic Clique Rescaling}, 
  year={2024},
  volume={46},
  number={12},
  pages={9780-9796},
  doi={10.1109/TPAMI.2024.3432099}}

@ARTICLE{10433795,
  author={Tu, Yunbin and Li, Liang and Su, Li and Zha, Zheng-Jun and Huang, Qingming},
  journal={IEEE Transactions on Pattern Analysis and Machine Intelligence}, 
  title={SMART: Syntax-Calibrated Multi-Aspect Relation Transformer for Change Captioning}, 
  year={2024},
  volume={46},
  number={7},
  pages={4926-4943},
  doi={10.1109/TPAMI.2024.3365104}}

@article{peng2025survey,
  title={A Survey on Fine-Grained Multimodal Large Language Models},
  author={Peng, Yuxin and Wang, Zishuo and Li, Geng and Zheng, Xiangtian and Yin, Sibo and He, Hulingxiao},
  journal={Authorea Preprints},
  year={2025},
  publisher={Authorea}
}

@article{ll5,
  title   = {LMDA: LLM-guided Marginal Distribution Alignment for Open-set Active Learning},
  author  = {Jingyi Tang and Li Liang and Beichen Zhang and Qingming Huang},
  journal = {Chinese Journal of Electronics},
  year    = {2026}
}

@inproceedings{zhang2024speaker2dubber,
  author       = {Zhedong Zhang and
                  Liang Li and
                  Gaoxiang Cong and
                  Haibing Yin and
                  Yuhan Gao and
                  Chenggang Yan and
                  Anton van den Hengel and
                  Yuankai Qi},
  title        = {From Speaker to Dubber: Movie Dubbing with Prosody and Duration Consistency
                  Learning},
  booktitle    = {Proceedings of the 32nd {ACM} International Conference on Multimedia,
                  2024},
  pages        = {7523--7532},
  year         = {2024},
}

@inproceedings{zhang2025prosody,
  title={Prosody-enhanced acoustic pre-training and acoustic-disentangled prosody adapting for movie dubbing},
  author={Zhang, Zhedong and Li, Liang and Yan, Chenggang and Liu, Chunshan and Van Den Hengel, Anton and Qi, Yuankai},
  booktitle={Proceedings of the IEEE/CVF Conference on Computer Vision and Pattern Recognition},
  pages={172--182},
  year={2025}
}

@inproceedings{cui2024stochastic,
  title={Stochastic context consistency reasoning for domain adaptive object detection},
  author={Cui, Yiming and Li, Liang and Zhang, Jiehua and Yan, Chenggang and Wang, Hongkui and Wang, Shuai and Jin, Heng and Wu, Li},
  booktitle={Proceedings of the 32nd ACM International Conference on Multimedia},
  pages={1331--1340},
  year={2024}
}

@inproceedings{cui2025debiased,
  title={Debiased Teacher for Day-to-Night Domain Adaptive Object Detection},
  author={Cui, Yiming and Li, Liang and Yin, Haibing and Gao, Yuhan and Sun, Yaoqi and Yan, Chenggang},
  booktitle={Proceedings of the IEEE/CVF International Conference on Computer Vision},
  pages={2577--2587},
  year={2025}
}

@inproceedings{zhao2026temporal,
  title={Temporal Calibrating and Distilling for Scene-Text Aware Text-Video Retrieval},
  author={Zhao, Zhiqian and Li, Liang and Shen, Lei and Sheng, Xichun and Sun, Yaoqi and Kang, Fang and Yan, Chenggang},
  booktitle={Proceedings of the AAAI Conference on Artificial Intelligence},
  volume={40},
  number={16},
  pages={13323--13331},
  year={2026}
}

@inproceedings{zhao2025heterogeneous,
  title={Heterogeneous Prompt-Guided Entity Inferring and Distilling for Scene-Text Aware Cross-Modal Retrieval},
  author={Zhao, Zhiqian and Li, Liang and Zhang, Jiehua and Sun, Yaoqi and Sheng, Xichun and Yin, Haibing and Jiang, Shaowei},
  booktitle={Proceedings of the AAAI Conference on Artificial Intelligence},
  volume={39},
  number={10},
  pages={10537--10545},
  year={2025}
}
}


\end{document}